\documentclass[10pt,conference]{IEEEtran}
\usepackage{amsmath,amsfonts,graphicx,hyperref}

\usepackage[natbib, bibencoding=utf8, citestyle=numeric, bibstyle=ieee, maxbibnames=999, maxcitenames=2, mincitenames=1, sortcites]{biblatex}
\bibliography{refs}

\usepackage{todonotes}
\usepackage[capitalize]{cleveref}
\usepackage{booktabs}
\usepackage{multirow}
\usepackage{graphicx}
\usepackage{subcaption}
\usepackage{xspace}

\usepackage[acronym, shortcuts, nohypertypes={acronym}]{glossaries}
\newacronym{AI}{AI}{artificial intelligence}
\newacronym{DL}{DL}{deep learning}
\newacronym{DNN}{DNN}{deep neural network}
\newacronym{RIR}{RIR}{room impulse response}
\newacronym{RL}{RL}{reinforcement learning}

\newcommand{\eg}{e.\,g.,}

\newcommand{\subplotwidththree}{0.24\textwidth}
\newcommand{\spacereducer}{-0cm}

\newacronym{AS}{AS}{\textit{AudioSet}}
\newacronym{ASC}{ASC}{\textit{acoustic scene classification}}
\newacronym{SCR}{SCR}{\textit{speech command recognition}}
\newacronym{MFCC}{MFCC}{\textit{Mel frequency cepstral coefficient}}
\newacronym{eGeMAPS}{eGeMAPS}{\textit{extended geneva minimalistic acoustic parameter set}}
\newacronym{ComParE}{ComParE}{\textit{computational paralinguistics challenge}}

\newacronym{CA}{CA}{\textit{computer audition}}
\newacronym{LEAF}{LEAF}{\textit{learnable frontend}}
\newacronym{TL}{TL}{\textit{transfer learning}}
\newacronym{UAR}{UAR}{\textit{unweighted average recall}}
\newacronym{CumAcc}{CumAcc}{\textit{cumulative accuracy}}
\newacronym{XAI}{XAI}{\textit{explainable artificial intelligence}}
\newacronym{SNR}{SNR}{\textit{signal to noise ratio}}
\newacronym{CV}{CV}{\textit{computer vision}}
\newacronym{AA}{AA}{\textit{audio analysis}}
\newacronym{ML}{ML}{\textit{machine learning}}
\newacronym{OOD}{OOD}{\textit{out of distribution}}
\newacronym{MIR}{MIR}{\textit{music information retrieval}}
\newacronym{DSP}{DSP}{\textit{digital signal processing}}
\newacronym{ASR}{ASR}{\textit{automatic speech recognition}}
\newacronym{SER}{SER}{\textit{speech emotion recognition}}
\newacronym{BaCE}{BalancedCE}{\textit{balanced cross-entropy}}
\newacronym{BCE}{BCE}{\textit{binary cross-entropy}}

\newacronym{HNR}{HNR}{\textit{harmony-to-noise-ratio}}
\newacronym{PCEN}{PCEN}{\textit{per-channel energy normalisation}}
\newacronym{CNN}{CNN}{\textit{convolutional neural network}}
\newacronym{LSTM}{LSTM}{\textit{long short-term memory}}
\newacronym{PEFT}{PEFT}{\textit{parameter efficient fine-tuning}}
\newacronym{LLM}{LLM}{\textit{large language model}}
\newacronym{BAD}{BAD}{\textit{bird activity detection}}

\newacronym{SoT}{SoT}{\textit{sound of things}}
\newacronym{S}{S}{\textit{speech}}
\newacronym{HS}{HS}{\textit{human sounds}}
\newacronym{A}{A}{\textit{animal}}
\newacronym{NS}{NS}{\textit{natural sounds}}

\newcommand{\wrt}{w.\,r.\,t.\xspace}

\begin{document}

\title{How Class Ontology and Data Scale Affect Audio Transfer Learning\\
\thanks{Identify applicable funding agency here. If none, delete this.}
}

\author{
  \textbf{Manuel Milling}$^{1,2,3}$, \textbf{Andreas Triantafyllopoulos}$^{1,2,3}$, \textbf{Alexander Gebhard}$^{1,2,3}$,\\\textbf{Simon Rampp}$^{1,2,3}$, \textbf{Bj\"orn W.\ Schuller}$^{1,2,3,4}$\\
  $^1$CHI -- Chair of Health Informatics, Technical University of Munich, Munich, Germany\\
  $^2$MCML -- Munich Center for Machine Learning, Munich, Germany\\
  $^3$MDSI -- Munich Data Science Institute, Munich, Germany\\
  $^4$GLAM -- Group on Language, Audio, \& Music, Imperial College, London, UK\\
  \texttt{manuel.milling@tum.de}
}


\maketitle

\begin{abstract}
Transfer learning is a crucial concept within deep learning that allows artificial neural networks to benefit from a large pre-training data basis when confronted with a task of limited data. 
Despite its ubiquitous use and clear benefits, there are still many open questions regarding the inner workings of transfer learning and, in particular, regarding the understanding of when and how well it works. 
To that extent, we perform a rigorous study focusing on audio-to-audio transfer learning, in which we pre-train various model states on (ontology-based) subsets of AudioSet and fine-tune them on three computer audition tasks, namely acoustic scene recognition, bird activity recognition, and speech command recognition.
We report that increasing the number of samples and classes in the pre-training data both have a positive impact on transfer learning. This is, however, generally surpassed by similarity between pre-training and the downstream task, which can lead the model to learn comparable features.
\end{abstract}

\section{Introduction}\label{sec:intro}
A key role in the success of \acp{DNN} is their ability to learn generalisable patterns from a vast amount of data and leverage them to new tasks or domains, a feat which can be summarised under the umbrella term \ac{TL}~\cite{zhuang2020comprehensive}.
While \ac{TL} encapsulates a large variety of related concepts, including few-shot~\cite{wang2020generalizing} and zero-shot learning~\cite{pourpanah2022review}, 
\ac{PEFT}~\cite{ding2023parameter, hu2022lora},
and fine-tuning,
its importance is now arguably higher than ever in the era of foundation models~\cite{bommasani2021opportunities}. 
Independent of the specifics, the core principle of \ac{TL} lies in the learning of powerful representations that generalise well across domains and therefore only need small levels of adaptation to perform well in a variety of downstream tasks. 

However, understanding the drivers leading to good representations remains an open research question.
An early, impactful attempt in the \ac{CV} domain was made by \citet{huh2016makes}.
The authors investigated ``what makes ImageNet good for transfer learning'' by testing how the amount of data, the granularity of labels, and the similarity of tasks seen during pre-training impact fine-tuning task performance.
They concluded that pre-training on ImageNet~\cite{deng2009imagenet} was primarily beneficial -- even with substantially reduced amounts of data.
Interestingly, supervised pre-training on a disjoint set of classes did not impair downstream task performance.
This contradicts the intuitive assumption that pre-training on more similar data and tasks can improve performance, as seen, for instance, in \citet{yosinski2014transferable}.

With the rise of auditory foundation models~\citep{Triantafyllopoulos25-CAF}, these questions become highly relevant for the domain of \ac{CA} as well.
To the present, very few studies systematically evaluate the properties of pre-training data that lead to good downstream generalisation for audio.
Examples for this are the comparison between image and audio data for pre-training in \citet{tomoya2020} or that of the role of task similarity in the context of \ac{SER}~\cite{triantafyllopoulos2021role}. 
Yet none of these studies covers a range of \ac{CA} tasks, which is a necessary prerequisite for understanding the drivers of \ac{TL} in the audio realm.

This is a considerable gap in present literature, especially since the audio domain can at times show distinct effects compared to \ac{CV} (\eg~\cite{milling2024bringing}) and does offer a similar and popular setup with many parallels -- namely, AudioSet~\cite{ds-audioset}. This dataset is of comparative size to ImageNet and commonly used to pre-train \acp{DNN} that are then fine-tuned for different \ac{CA} tasks~\cite{kong2020panns}.
This gap is important as auditory foundation models are becoming more and more pervasive in the present state-of-the-art~\citep{Triantafyllopoulos25-CAF}.
These models are, almost without exception, all pre-trained on AudioSet as the largest available (labelled) corpus of general audio data.
The underlying assumption is that including the largest available dataset, which also features the broadest available ontology, is the optimal strategy for pre-training.
The corollary of this assumption is that future iterations of pre-training datasets should have similar characteristics -- substantial amounts of data derived from a broad ontology.

Yet the findings of \citet{huh2016makes} contradict this assumption.
Beyond that, several works have found that pre-training audio models on ImageNet data can lead to outstanding model performance~\citep{ren2018deep, amiriparian17_interspeech, gwardys2014deep, gong2021psla, guzhov2021esresnet} -- in some cases, even preferable to pre-training on AudioSet data~\citep{Wagner24-ASE}.
This surprising finding underscores how basic and intuitive assumptions do not necessarily hold after rigorous evaluation.
More importantly, this raises the question of how to move forward.
Given that general audio analysis is not yet a solved problem, the community will inevitably need to collect more data (a problem exacerbated by the inevitable decline of AudioSet as more and more videos are removed from YouTube over time).
Our contribution aims to elucidate the factors that drive improved transfer performance from a dataset perspective.



To address the presented research gap, we present the following contributions:
\begin{itemize}
    \item we provide a set of model states that are pre-trained on various subsets of AudioSet,
    \item we fine-tune these model-states on a variety of \ac{CA} tasks, namely \ac{ASC}, \ac{BAD} and \ac{SCR},
    \item we analyse how task similarity, the number of samples, as well as the number of classes in the pre-training data impact effects of \ac{TL} and similarity of pre-trained model states. 
\end{itemize}


\vspace{\spacereducer}
\section{Experimental Design}

\subsection{Datasets}\label{sec:data}
We select a variety of datasets to investigate the importance of (ontological) task and data similarity between pre-training and fine-tuning datasets for the success of audio \ac{TL}. \\

\begin{figure}[t]
  \centering

     \centering
     \includegraphics[width=0.75\columnwidth]{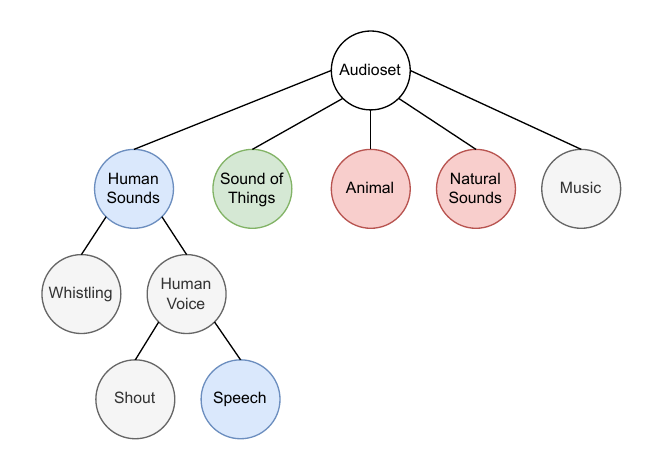}
  \caption{
   Excerpt of the AudioSet ontology, including specific nodes being considered in the selection of pre-training data. We colour the nodes whose branches explicitly serve as a data basis for pre-training schematically, based on whether they are focused on sounds related to humans (blue), nature (red), or mechanical things (green).  
  }
  \label{fig:AudioSetOntology}
\end{figure}

\textbf{\Acf{AS}}: \ac{AS} is one of the largest, publicly available, labelled audio datasets~\cite{ds-audioset} and is thus a very popular choice for the pre-training of audio models. 
It represents a multi-label audio tagging dataset with 10\,s segments labelled with multiple tags. 
The data contains a large variety of often overlapping audio events with the labels being designed according to a rich ontology, a small part of which is illustrated in~\cref{fig:AudioSetOntology}. 
This makes \ac{AS} a particularly useful dataset to investigate effects of pre-training as we can construct large subsets of the data and train them only on certain categories of labels. \Cref{tab:exp_tl_audioset-subsets} contains an overview of the considered subsets of \ac{AS} for pre-training, which are representing (combinations of) sounds with human (especially speech), nature-related (animals, wind, etc.) or mechanical (e.g., vehicles) origin. 
Concretely, for the described subsets, we only consider labels that are within the branch of the highlighted node and only include samples that have at least one of these labels.
Other labels are ignored for the multi-label classification problem but some of them are present as background noise in some recordings, i.\,e., we did not exclude segments that were labelled with sounds beyond each label subset; we only excluded segments which did not have any of the selected labels in each subset. 
This selection constitutes the different number of samples and classes denoted in \Cref{tab:exp_tl_audioset-subsets}.
In addition to the ontology-based subsets in \Cref{tab:exp_tl_audioset-subsets}, we also perform pre-training on randomly sampled subsets of the full \ac{AS} corpus with all available positive labels. 
\\

\textbf{\Acf{ASC}}: Our first fine-tuning task is the 2020 version of the DCASE \ac{ASC} challenge (task 1A)~\cite{ds-dcase-2020}. In order to assign 10\,s long audios to one of ten classes, the model needs to perceive a variety of audio events, both from mechanical (as a moving tram) and natural sources (as birds chirping in a park).
\textit{Human sounds} in general and \textit{speech}, in particular, should play a lesser role within the data, as many instances have been removed to protect privacy, while \textit{sounds of things}, \textit{animal} sounds, and \textit{natural sounds} should be abundant.\\

\textbf{\Acf{BAD}}: Our second fine-tuning task is \ac{BAD}, which was introduced as part of the DCASE2018 challenge~\cite{Stowell18-AAD}, the model needs to decide whether each segment contains bird vocalisations. The data should be closely related to the \textit{animal} and \textit{natural sounds} branches of the \ac{AS} ontology.\\

\textbf{\Acf{SCR}}: The final fine-tuning task, \ac{SCR}, is a simple implementation of an \ac{ASR} task, in which short recordings of around 1\,s need to be assigned to one of 35 key words~\cite{warden2018speech}. The task thus has a particular connection to the \textit{speech} branch of the \ac{AS} ontology, even though \ac{AS} does not provide labels based on linguistic content.


\subsection{Experiments}
We conduct transfer learning experiments in the form of pre-training and fine-tuning without frozen weights, i.\,e., we first randomly initialise the parameters of the model architecture and initially train them on the pre-training data (multi-label classification). In the second step, we exchange the classification layer with a new randomly initialised classification layer matching the number of classes of the downstream fine-tuning task and continue the training of all parameters on the fine-tuning data. An overview of the training setting and hyperparameters employed, both during pre-training and fine-tuning stages, is provided in \Cref{tab:exp_tl_parameter_table}. 

Pre-training experiments are performed on all subsets from \Cref{tab:exp_tl_audioset-subsets}, as well as randomly sampled subsets of \ac{AS} with a fraction of 30\,\%, 10\,\%, 3\,\%, and 1\,\%, respectively, leading to a total of 13 pre-training experiments -- only with a single random seed.
Note that with a constant number of pre-training epochs and batch size across data bases for pre-training, the number of optimisation steps decreases with decreasing data set size. 
We decided for this implementation, as 1) generalisation saturates quicker with smaller datasets~\cite{kaplan2020scaling} and we aim to avoid negative effects of transfer learning on heavily overfitted model states~\cite{deng2023towards, springer2025overtrained}, 2) the resource consumption stays more manageable for the already heavy experiments with the pre-training on the full AudioSet requiring 78\,h wall time in our setup.  
Fine-tuning is performed for each combination of pre-training state, fine-tuning dataset, and random seed, leading to a total of 126 fine-tuning experiments with roughly 30\,min to 60\,min wall time per experiment, depending on the fine-tuning task.

We choose to focus our experiments on the singular, well-established architecture CNN10~\cite{kong2020panns}, given the already extensive amount of pre-training and fine-tuning experiments. 
Recordings are resampled to 16\,kHz and Mel-spectrograms are extracted as specified in \citet{kong2020panns} to serve as inputs to the \ac{CNN}.
\Ac{BaCE} loss is employed to counter the effects of class imbalance during fine-tuning. 
The fine-tuning task performance is reported in terms of \ac{UAR} on the respective test dataset with the best-performing model state on the validation data and is averaged across the three random seeds.
Our experiments and analysis are based on the PyTorch-based autrainer toolkit~\cite{rampp2024autrainer}, and we follow the respective defined splits for the training, development, and testing sets therein. The code as well as (pre-)trained model states are available under\footnote{
\url{https://github.com/millinma/OntoTLAudio}
}. 
Training was performed on NVIDIA A100 GPUs with 80\,GB VRAM.
\begin{table}[t]
  \caption{Subsets of AudioSet considered for pre-training, including information about the corresponding ontology branches, number of training samples, positive classes, and subset-specific performance on the pre-training task. Unweighted average F1 (UAF1)-score is based on a threshold of 0.5. Abbreviations: \acf{S}, \acf{HS}, \acf{SoT}, \acf{A}, \acf{NS}. \label{tab:exp_tl_audioset-subsets}
}
  \centering
  \footnotesize
  \begin{tabular}{lrrrr}
    \toprule
    \textbf{Abbreviation} & \textbf{\# data points} & \textbf{Branches} & \textbf{\# classes} & \textbf{UAF1}\\
    \midrule
    AS & 1\,716\,492 & all & 527 & 0.194\\ 
    AS-speech & 836\,623 & \acs{S} & 8 & 0.222\\
    AS-humansounds & 910\,601 & \acs{HS} & 72 & 0.230\\
    AS-things & 246\,927 & \acs{SoT} & 167 & 0.309\\
    AS-animal & 81\,237 & \acs{A} & 65 & 0.442\\
    AS-naturalsounds & 31\,310 & \acs{NS} & 18 & 0.493 \\
    AS-nature & 110\,144 & \acs{A}, \acs{NS} & 82 & 0.425\\
    AS-nonhuman & 344\,532 & \acs{A}, \acs{NS}, \acs{SoT} & 256 & 0.319\\
    AS-naturalsources & 981\,575 & \acs{HS}, \acs{A}, \acs{NS} & 154 & 0.292\\

    \bottomrule
  \end{tabular}
\end{table}

\begin{table}[t]
  \caption{Overview of the training setup for the pre-training and fine-tuning stage of transfer learning experiments. Abbreviations: \acf{AS} \acf{ASC}, \acf{SCR}, \acf{BAD}, \acf{BCE}, \acf{BaCE}. \label{tab:exp_tl_parameter_table}
}
  \centering
  \footnotesize
  \begin{tabular}{lrr}
    \toprule
        \textbf{Training Setting} & \textbf{Pre-training} & \textbf{Fine-Tuning} \\
    \midrule
    Datasets & \acs{AS}(-subset) & \acs{ASC}, \acs{BAD}, \acs{SCR} \\

    Architecture & \multicolumn{2}{c}{CNN10}\\
    Optimiser & \multicolumn{2}{c}{Adam}\\
    Learning Rate & \multicolumn{2}{c}{$10^{-3}$} \\
    Batch Size & 128 & 32 \\
    Epochs & 30 & 50\\
    Loss & \acs{BCE} & \acs{BaCE} \\
    Random Seed & 1 & 1, 2, 3\\
    \bottomrule
  \end{tabular}
\end{table}

\vspace{\spacereducer}
\section{Results \& Discussion}
\subsection{Pre-training Performance}
Despite the limited comparability of pre-training performance across the subsets of AudioSet in \Cref{tab:exp_tl_audioset-subsets} due to differences in the data basis, we note a few insights:
The \emph{AS-speech} subset, as well as all subsets that include the \textit{speech} node, or its parent node \textit{human sounds}, show a particularly low performance, despite a -- generally advantageous -- high number of samples and a low number of classes. 
One possible reason for this might be the inconsistent labelling convention within AudioSet that does not always include higher ontology nodes in the segment labels (e.\,g., samples labelled as ``male speech'' are not always labelled as ``human sounds'' too).
\vspace{\spacereducer}
\subsection{Data Set Size and Ontology}
\begin{figure}[t]
  \centering
     \begin{subfigure}[b]{\subplotwidththree}
         \centering
         \includegraphics[width=\textwidth]{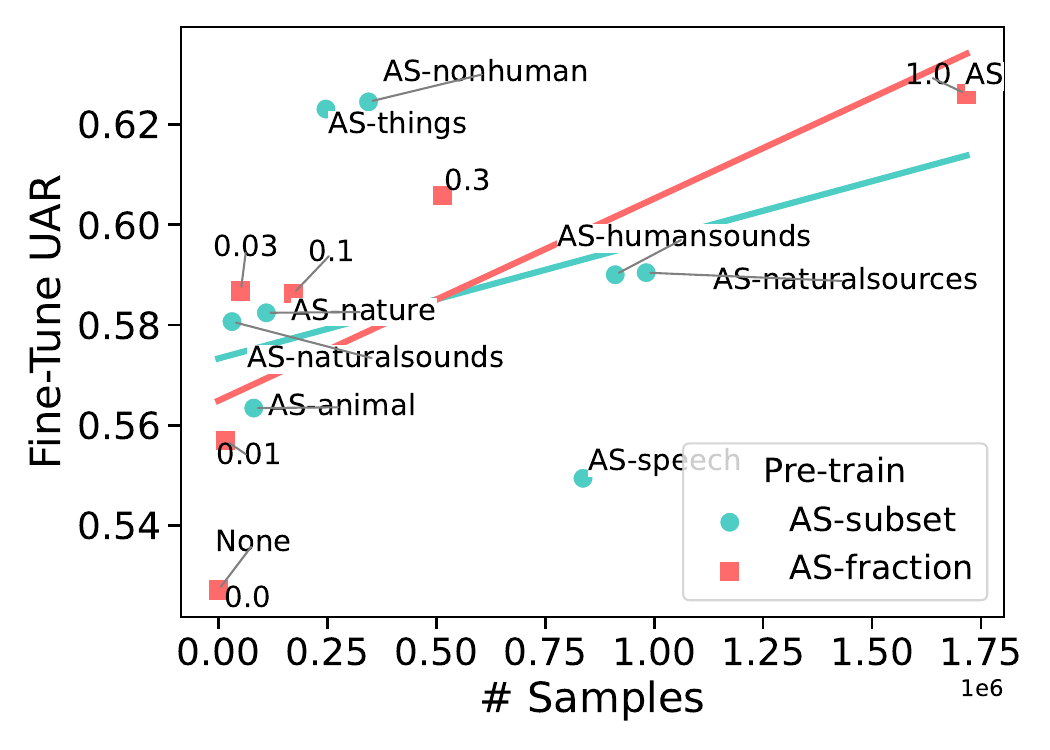}
         \caption{ASC}
     \end{subfigure}
     \begin{subfigure}[b]{\subplotwidththree}
         \centering
         \includegraphics[width=\textwidth]{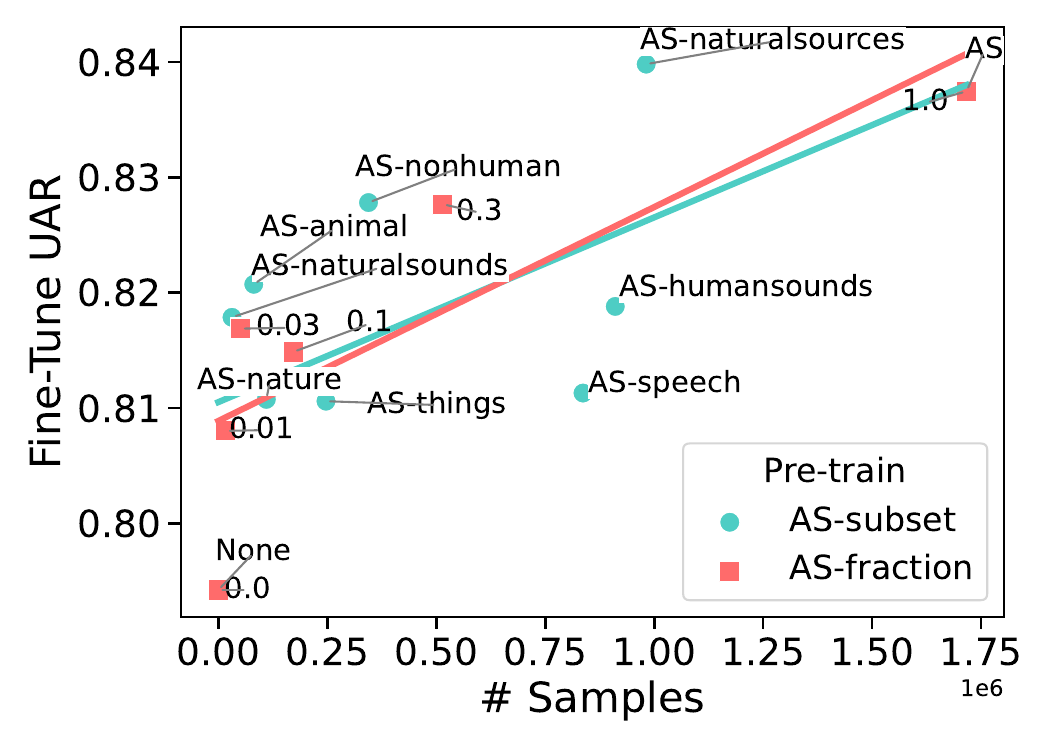}
         \caption{BAD}
     \end{subfigure}
     \begin{subfigure}[b]{\subplotwidththree}
         \centering
         \includegraphics[width=\textwidth]{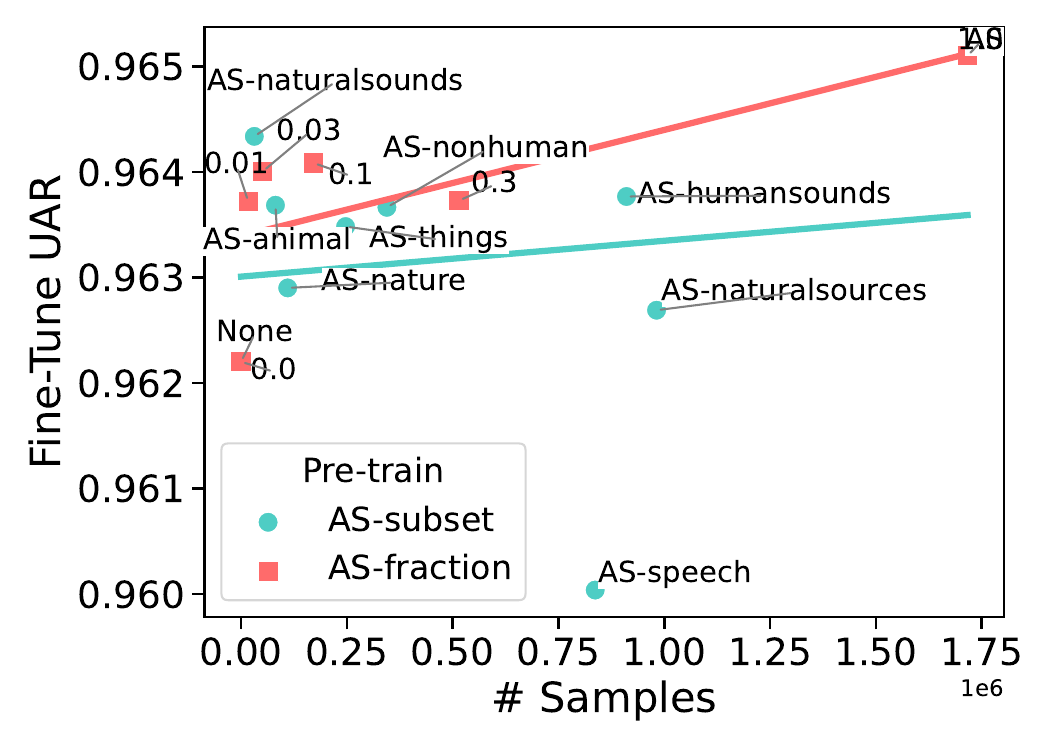}
         \caption{SCR\label{subfig:SCR_samples}}
     \end{subfigure}
  \caption{
   Performance of fine-tuning experiments on the three datasets \acf{ASC}, \acf{BAD},  and \acf{SCR} \wrt the number of pre-training samples. We consider different pre-training states trained on randomly sampled (orange) and ontology-based (blue) subsets of \acf{AS}. Results are averaged across three random seeds. Note the difference in performance scale across the fine-tuning tasks. 
  }
  \label{fig:exp_tl_subset_samples}
\end{figure}

The subfigures in \Cref{fig:exp_tl_subset_samples} display the performance for the respective fine-tuning tasks \wrt the number of pre-training samples and the pre-training task. 
We differentiate between the randomly sampled subsets of the full \ac{AS} (orange) and the ontology-specific subsets of \ac{AS} (blue). 

We observe a clear positive correlation between the number of samples in the pre-training data set and downstream task performance in all cases, which is especially prominent for the randomly sampled subsets of \ac{AS}. This clearly emphasises the important role of the number of pre-training samples in achieving positive transfer. 
Importantly, almost all pre-trained model states obtain better performance over the baseline (``None''), consisting of a randomly initialised state.
This indicates that positive transfer occurs despite the amount of pre-training data being small -- even comparable to the amount of data available for the downstream task.

However, the correlation between the number of samples in pre-training and fine-tuning consistently decreases across the board when we consider the ontology-based subsets compared to the randomly sampled ones. 
While the correlation stays positive, it is much less prominent, indicating that the ontology-based selection of positive labels has important implications. 
For instance, for both the \gls{ASC} and \gls{BAD} datasets, pre-training on some subsets can lead to similar and even better \ac{TL} effects than on the full AudioSet.

In this context, the importance of similarity between pre-training and fine-tuning task becomes apparent:
for \ac{ASC}, it is important to analyse sound events, amongst which mechanical sounds (\eg class \textit{tram}) and environmental sounds (\eg class \textit{park}) play a more important role than speech signals.
In line with this assumption, some positive outliers (\wrt the regression line) in~\Cref{fig:exp_tl_subset_samples} can be observed for AudioSet subsets that include the branches 
\emph{natural sounds}, and, most notably, \emph{sounds of things}. 
Pre-training branches that do not include the \emph{sounds of things} node, on the other hand, show a lower \ac{TL} benefit for \ac{ASC}.
The highest \ac{TL} effects for \ac{BAD} can be observed with pre-training on \emph{AS-naturalsources}, which includes all considered nodes, except \textit{sounds of things}, with results even surpassing the performance after pre-training on the full AudioSet. Pre-training on the \emph{sounds of things} branch, on the other hand, does not show any particular benefit.

We do note some counter-intuitive results, most prominently, the poorer performance on the \gls{SCR} task with pre-training on the \emph{speech} branch.
We note that performance on \gls{SCR} is already very high ($>.96$) without pre-training, a fact which clearly limits the meaningfulness of conclusions drawn about the impacts of \gls{TL} as performance has already saturated. 
We nevertheless include the results in the discussion as the selected \ac{SCR} task serves as a common dataset for \ac{CA} benchmarks~\cite{meng23c_interspeech, zeghidour2021leaf, gong21b_interspeech}.
Another example is that the performance on the \gls{BAD} task improves when we add the generally unrelated \emph{sounds of things} branch to the \emph{AS-nature} subset -- thus constituting the \emph{AS-nonhuman} subset.

While some of these inexplicable differences may be caused solely due to statistical chance, we have aimed to mitigate random effects by averaging over multiple seeds.
Rather, we interpret these observations as further indications that we cannot easily intuit about the success of a particular pre-training subset based on our subjective understanding of the data.

\vspace{\spacereducer}
\subsection{Class Granularity and Ontology}
\begin{figure}[t]
  \centering
     \begin{subfigure}[b]{\subplotwidththree}
         \centering
         \includegraphics[width=\textwidth]{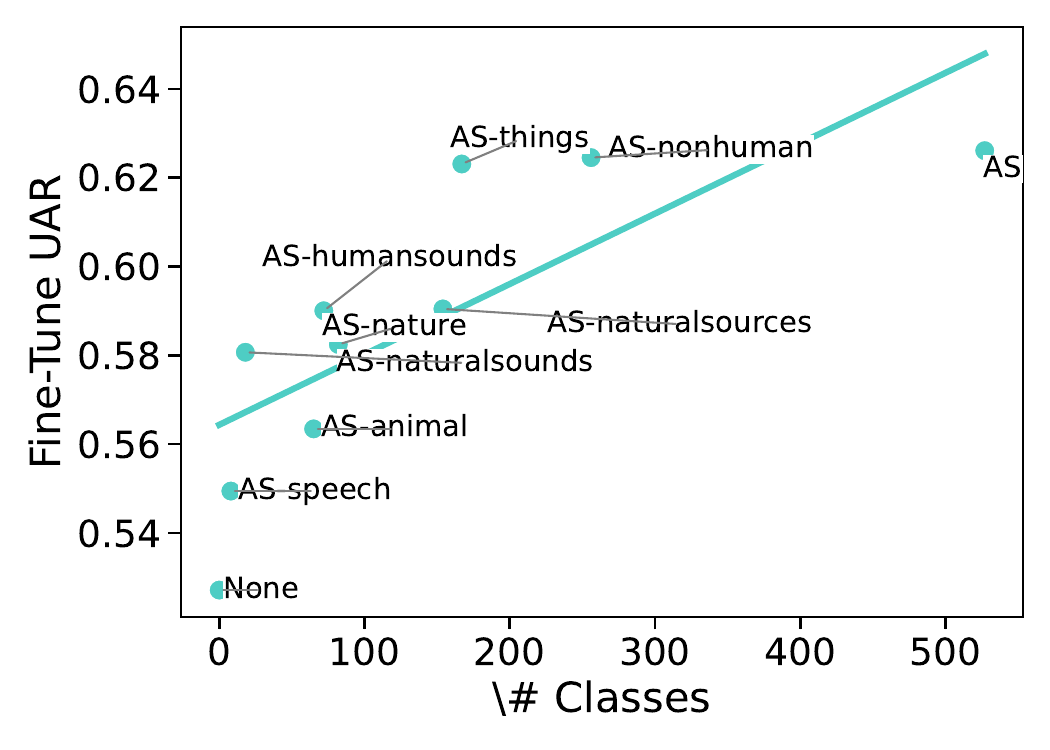}
         \caption{ASC}
     \end{subfigure}
     \begin{subfigure}[b]{\subplotwidththree}
         \centering
         \includegraphics[width=\textwidth]{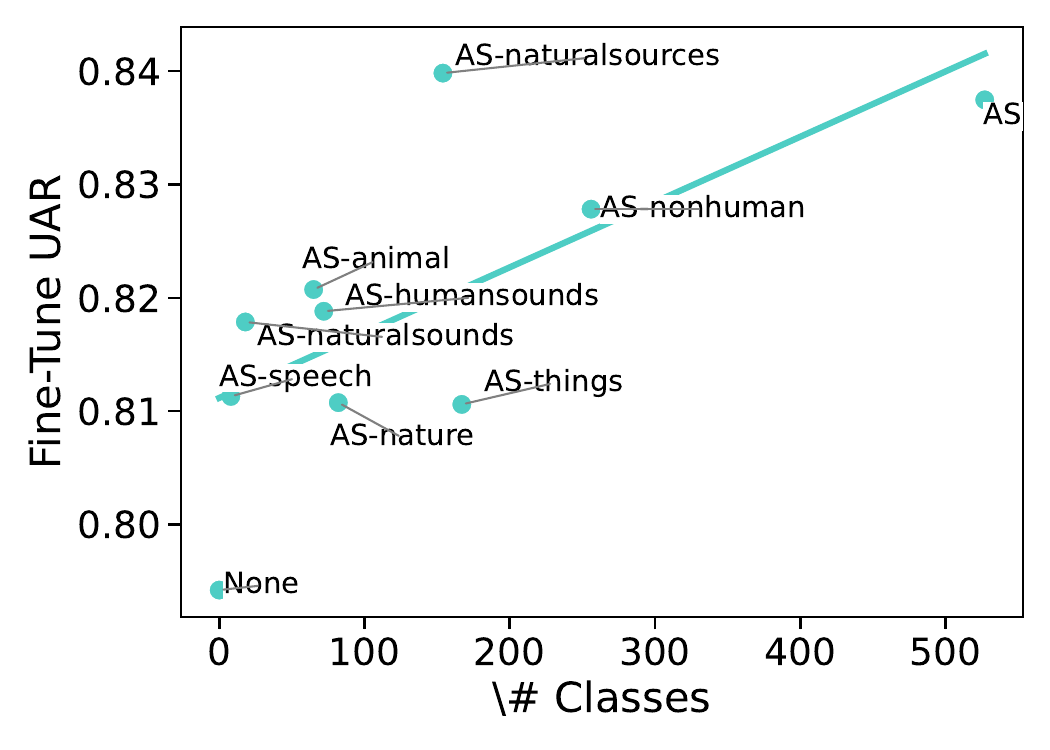}
         \caption{BAD}
     \end{subfigure}
     \begin{subfigure}[b]{\subplotwidththree}
         \centering
         \includegraphics[width=\textwidth]{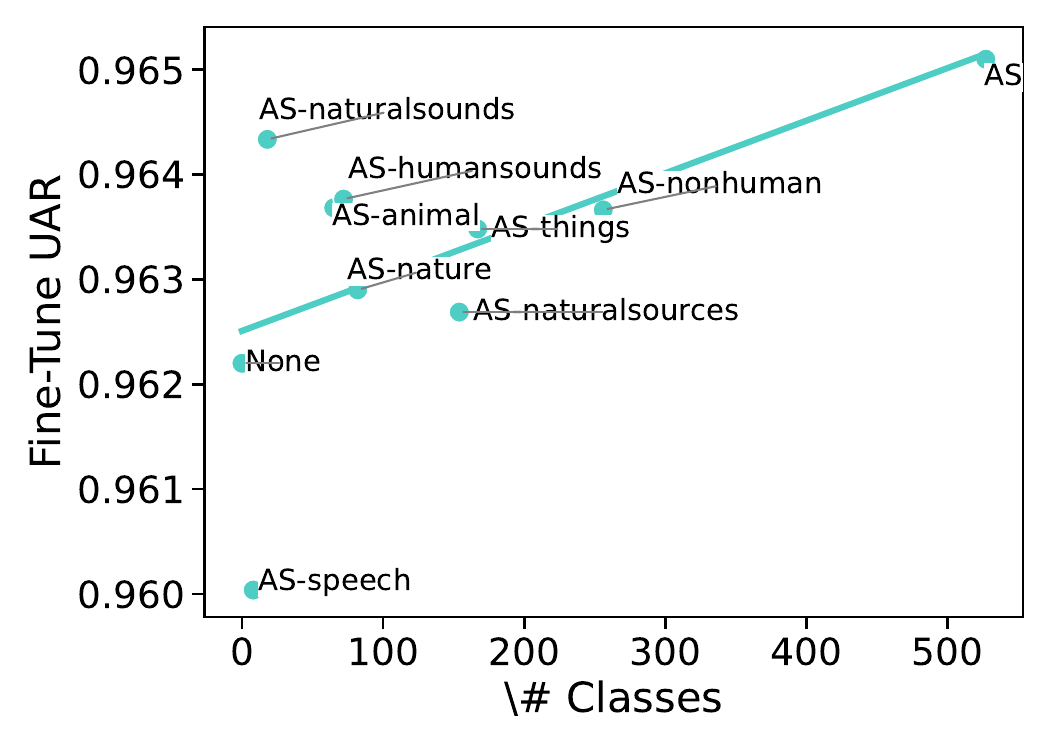}
         \caption{SCR\label{subfig:SCR_classes}}
     \end{subfigure}
  \caption{
   Performance of fine-tuning experiments on the three datasets \acf{ASC}, \acf{BAD} and \acf{SCR} \wrt the number of pre-training classes. We consider different pre-training states trained on ontology-based subsets of \acf{AS}. Results are averaged across three random seeds. Note the difference in performance scale across the fine-tuning tasks.
  }
  \label{fig:exp_tl_subset_classes}
\end{figure}

For the number of classes in the pre-training data, we report once more a positive correlation \wrt downstream task performance in~\Cref{fig:exp_tl_subset_classes}, this time even higher compared to the number of samples.
A higher granularity and variety of labels thus seems to be important for successfully pre-training a model state, even more so than the number of samples. 
This may also explain the low performance of \textit{AS-speech} state, which is based on a large number of pre-training samples but a low number of pre-training classes.
Outliers of ``overperforming'' and ``underperforming'' pre-training states \wrt the number of pre-training classes
can once more, to a certain degree, be attributed to the resemblance of tasks.

\subsection{Similarity of Representations}
\begin{figure}[t]
  \centering

     \centering
     \includegraphics[width=0.9\columnwidth]{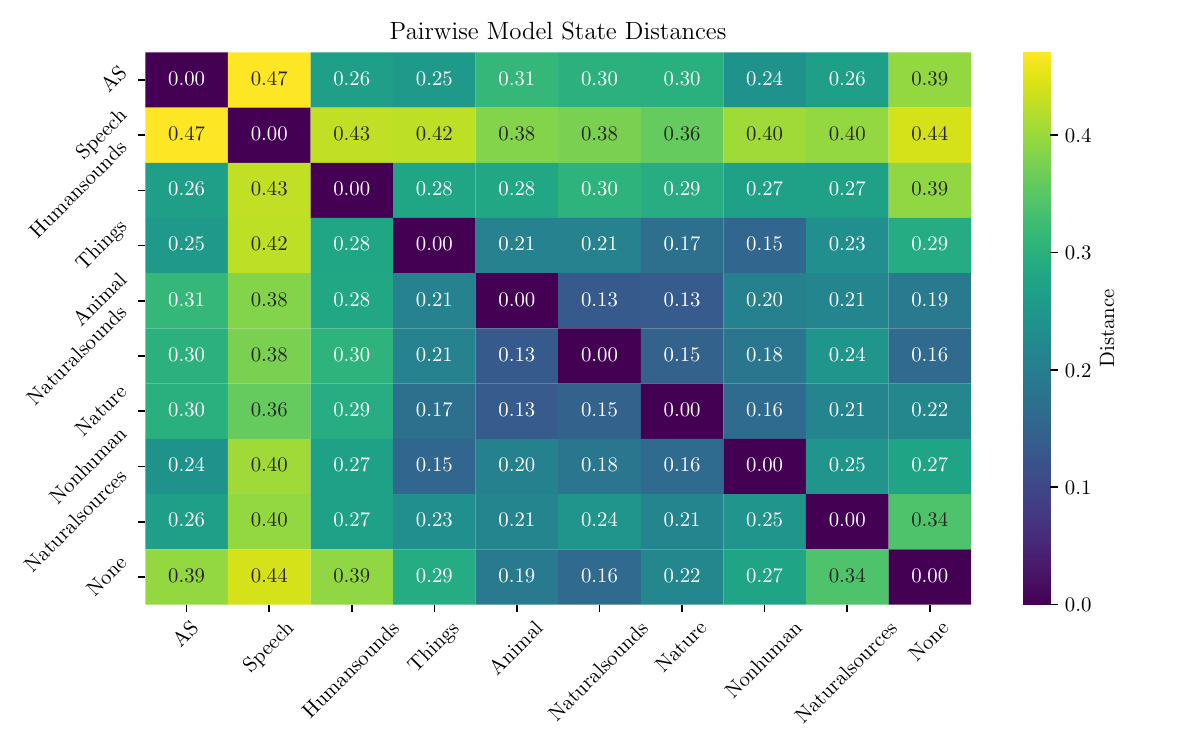}
  \caption{
     Pair-wise cosine distance between the first convolutional layer of model states, pre-trained on a different subset of AudioSet.  
  }
  \label{fig:pairwise_cosinedistance}
\end{figure}

To better understand how the individual pre-trained model states relate to each other, \Cref{fig:pairwise_cosinedistance} displays the pairwise cosine distance of the first convolutional layer of the CNN10 architectures across the different ontological pre-training bases. We choose to only focus on the first convolutional layer, as it is said to represent the most basic, fundamental features for a task, at least within \acp{CNN} of the \ac{CV} domain~\cite{zeiler2014visualizing}.
We observe a particularly low cosine distance amongst the subsets that are thematically close, such as \textit{Animal} and \textit{Naturalsounds} or amongst those with shared data basis, such as their combination \textit{Nature}, or the set of \textit{Nonhuman} and the included set of \textit{Things}. Differences from the randomly initialised state (\textit{None}) show a tendency to increase with increased size of pre-training data, which is likely connected to the correspondingly increased training time. Finally, the \textit{Speech} subset shows a particularly high distance to all other states, indicating dissimilar representations, which are likely the cause of the particularly bad fine-tuning performance. 
Overall, \Cref{fig:pairwise_cosinedistance} suggests that the pre-training data has a large impact on the learned representations during pre-training, at least at the fundamental feature layers. This emphasises the importance of not only an adequate pre-training data basis for \ac{TL} but also efforts towards the learning of task-specific features, as for instance demonstrated by the \ac{LEAF}~\cite{zeghidour2021leaf}.    

\vspace{\spacereducer}
\subsection{Limitations}
We note several limitations to our work, which are largely caused by computational limitations that prevented us from going beyond the already notable resource consumption for the presented study. First, our experiments are limited to a single \gls{CNN} architecture, a specific selection of hyperparameters and, in particular, a limited number of epochs. 
Furthermore, despite the success of \ac{AS} for transfer learning in the audio domain and its insightful ontology, some limitations remain between the relation of the \ac{AS} audio tagging task with specific classes and the corresponding downstream tasks.
In particular, as we did not exclude samples labelled with other tags, but only discarded those tags as target, the network has been exposed to data from them.
Nevertheless, we expect the model to discard such additional sources as background noise and not learn any relevant features for their prediction beyond simple, superficial statistics.

\vspace{\spacereducer}
\section{Conclusions}
In this contribution, we explored what makes audio-to-audio transfer learning work 
by pre-training on various subsets of AudioSet and fine-tuning on a set of selected \ac{CA} tasks. 
We discovered that both a larger number of classes and of samples in the pre-training have a positive impact on fine-tuning performance, with the impact of the former being generally higher. 
Furthermore, we observed that a similarity between the pre-training classes and the fine-tuning task often outweighs the aforementioned effect, likely making this the most crucial aspect in audio transfer learning. 
This finding is further underlined by representational resemblance between the pre-trained model state with ontologically related data bases.
Similar to the questions we pose in this contribution within the \ac{CA} domain, we aim to investigate in future work when and why the non-obvious transfer from the \ac{CV} to the \ac{CA} domain works. 

\section*{Acknowledgements}
This work was partially funded by the DFG's Reinhart Koselleck project No.\ 442218748 (AUDI0NOMOUS). We also acknowledge support from the NVIDIA Academic Grant Program.
\section{References}
\printbibliography[heading=none]

\end{document}